\documentclass[runningheads]{llncs}
\usepackage[T1]{fontenc}
\usepackage{graphicx}
\usepackage{cite}
\begin{document}
\title{A Consensus-Driven Multi-LLM Pipeline for Missing-Person Investigations}
%
%
\author{Joshua Castillo \and
Ravi Mukkamala$^*$\orcidID{0000-0001-6323-9789}}
\authorrunning{J. Castillo and R. Mukkamala}
%
\institute{Old Dominion University, Norfolk VA 23529, USA\\
\email{\{jcast046,rmukkama\}@odu.edu}}
\maketitle              
\begin{abstract}
The first 72 hours of a missing-person investigation are critical for successful recovery. Guardian is an end-to-end system designed to support missing-child investigation and early search planning. This paper presents the Guardian LLM Pipeline, a multi-model system in which LLMs are used for intelligent information extraction and processing related to missing-person search operations. The pipeline coordinates end-to-end execution across task-specialized LLM models and invokes a consensus LLM engine that compares multiple model outputs and resolves disagreements. The pipeline is further strengthened by QLoRA-based fine-tuning, using curated datasets. The presented design aligns with prior work on weak supervision and LLM-assisted annotation, emphasizing conservative, auditable use of LLMs as structured extractors and labelers rather than unconstrained end-to-end decision makers.

\end{abstract}
Research Track 001: APPLIED ARTIFICIAL INTELLIGENCE AND DATA SCIENCE
\section{Introduction}
Missing-child search planning is a complex, multidisciplinary process that requires the coordinated integration of information, expertise, and resources from multiple stakeholders. Effective search operations typically synthesize heterogeneous inputs, including last-known-position estimates, environmental and terrain data, weather conditions, sensor observations, and behavioral or mobility models of the missing subject \cite{ICRC2025}. These inputs are contributed by search-and-rescue (SAR) coordinators, field teams, subject-matter experts, data analysts, and increasingly by computational decision-support systems. The effectiveness of a missing-search operation therefore depends not only on the accuracy of individual data sources, but on the structured fusion of multi-party inputs to support timely, informed decision-making under uncertainty \cite{RuizReyes2025}.

Missing-child investigations begin with incomplete, rapidly evolving information and severe time constraints, particularly during the first 72 hours. Traditional early-stage search planning relies heavily on human judgment, coarse heuristics, and manual fusion of heterogeneous sources such as narrative reports, PDFs, public tips, transit data, and maps \cite{FBI2014}. In practice, the core challenge is not simply to predict a single location, but to produce calibrated uncertainty and actionable search products—such as prioritized regions and time-dependent likelihood surfaces—under extreme data sparsity \cite{RuizReyes2025, Solaiman2022}.

\begin{figure}
\includegraphics[width=\textwidth]{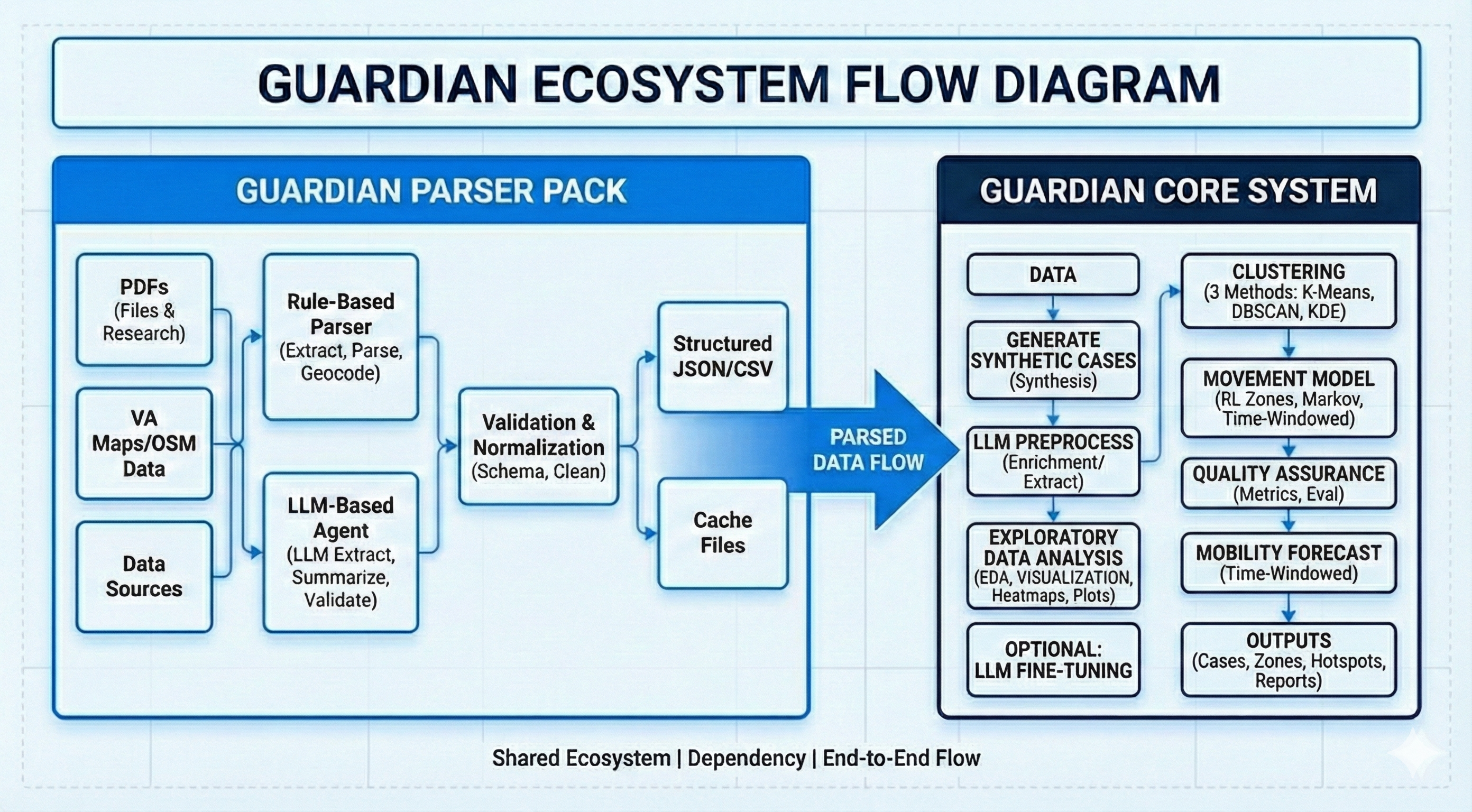}
\caption{Guardian System Architecture with two distinct but interconnected systems} \label{fig1}
\end{figure}

Guardian addresses these challenges through an end-to-end decision-support pipeline that converts raw, unstructured case documents into probabilistic search surfaces over a geographic grid and a set of human-interpretable artifacts, including ranked sectors, hotspots, and containment rings for 24-, 48-, and 72-hour horizons. Guardian is organized as a two-stage system (Figure 1). In Stage 1, the data preprocessing phase (Guardian Parser Pack), the system ingests heterogeneous raw inputs, normalizes and validates extracted fields, and enriches cases with external contextual data. In Stage 2, the analysis and evaluation phase (Guardian Core), the system performs structured validation, case generation, LLM-based processing with consensus, clustering and hotspot formation, probabilistic forecasting of ring-and-likelihood zones, search-plan generation, and plan evaluation. All outputs are designed to be auditable and consumable by investigators without requiring exposure to internal model mechanics.


In this paper, we focus on Guardian Pipeline, an LLM pipeline for information extraction and processing of the Guardian Core system. Given the time-sensitive nature of the application, rather than treating any single model as authoritative, Guardian treats each model as a fallible expert and routes all generated (predicted) information through a centralized, multi-model consensus layer. Thus, reliability is framed as a systems property. In Guardian, reliability is not treated as a single scalar score but as an operational property of the pipeline. More specifically, reliability refers to the degree to which the system produces valid, correct, and consistent outputs under model disagreement, malformed generations, and partial failures. In this paper, that notion includes structural correctness (for example, parseable and schema-aligned outputs), factual correctness relative to available ground truth, and cross-model consistency after normalization, repair, and consensus.
The pipeline produces concise investigator summaries, schema-aligned extractions, and weak (noisy or probabilistic) labels  that remain traceable to the underlying narrative and are suitable for downstream integration with hotspot detection, mobility forecasting, and geospatial planning components, which themselves demand spatially meaningful evaluation beyond generic accuracy metrics \cite{Aggarwal2015,Jiang2022,Lyu2025}.

The remainder of the paper is organized as follows. Section 2 summarizes related work. Section 3 provides details of different components of the proposed LLM consensus architecture. Section 4 describes the LLM promptings and their governance in this system. Section 5 provides a brief summary of the role of QLoRA in fine-tuning the models. In section 6, details of a qualitative evaluation of the system are provided. Section 7 has a discussion on the overall system along with its limitations. Finally, section 8 provides a summary and plans for future work.
\section{Related Work} 
Guardian draws methodological motivation from four intersecting research areas: missing-person decision support systems, unstructured document understanding, weak  (noisy, incomplete, or probabilistic) supervision and scalable labeling, and mobility-oriented geospatial modeling. 

Work in missing-person analytics and search optimization highlights the importance of fusing diverse data streams and generating actionable prioritizations rather than raw predictions \cite{Solaiman2022, RuizReyes2025}. Complementary literature in search and rescue demonstrates that probabilistic mapping and agent-based reasoning can convert sparse observations into interpretable density estimates \cite{Hashimoto2022, Ewers2024}, while GIS-based mobility algorithms formalize how movement patterns can be leveraged for planning \cite{Papic2024}.

For unstructured document analysis, surveys emphasize robust pipelines that normalize messy multi-format inputs into structured, analyzable representations, aligning with Guardian’s emphasis on schema-first processing and validation \cite{Mahadevkar2024}. In narrative-to-structure transformation, earlier work on entity extraction from police reports demonstrates the long-standing value of converting noisy narrative text into discrete entities and relations as a foundation for downstream analysis \cite{Chau2002}. Our Guardian system extends this direction by using LLMs for structured extraction while constraining outputs through schema validation and consensus-based reliability controls.

Research on weak supervision and label generation, where scalable training data is created by combining multiple noisy signals rather than relying solely on manual annotation \cite{Ratner2017}, is the primary motivation for us to employ it in the Guardian system. Recent work demonstrates that language models themselves can supply supervision signals for text classification, enabling learning under limited labeled data \cite{Zeng2022}. Within Guardian, LLMs are positioned as controlled labelers whose outputs are audited, cross-compared, and merged through consensus, consistent with findings that LLM-based annotation is particularly valuable when applied conservatively and validated \cite{Chen2024}. Finally, Guardian’s broader geospatial setting is informed by mobility forecasting literature and the recognition that spatial reasoning must be evaluated with spatially meaningful metrics, not solely generic accuracy measures \cite{Jiang2022, Lyu2025}.

Because missing-person workflows implicate sensitive data, Guardian’s design also aligns with ethical guidance advocating transparency, accountability, and risk-aware deployment of AI, especially where privacy and harm are salient \cite{Floridi2019, ICRC2025,Budowle2024PrivacyFGG}. Where synthetic data is employed for development and training, prior work highlights both its utility and the need to manage domain shift and privacy considerations \cite{Nikolenko2019,Sun2023SyntheticHealthData}.

\section{System Architecture}

Guardian's LLM pipeline (Figure 2) begins with an entry module (Case Narrative) that is a primary entry point for end-to-end execution across multiple cases. This module is responsible for loading cases, constructing narratives, initializing concurrency controls, and calling the pipeline in either a stage-by-stage mode or a case-by-case mode. 

Stage-by-stage processing is a throughput-oriented strategy in which the system runs all cases through the summarization stage, then all cases through extraction, and then all cases through weak labeling. This reduces repeated initialization overhead, improves cache locality, and enables more consistent resource utilization, a design objective aligned with classic algorithmic and systems concerns about batching and efficient scheduling \cite{Cormen2022}. 

Case-by-case processing, by contrast, is retained for debugging and interactive analysis where immediate end-to-end visibility is more valuable than throughput.

\begin{figure}
\includegraphics[width=\textwidth]{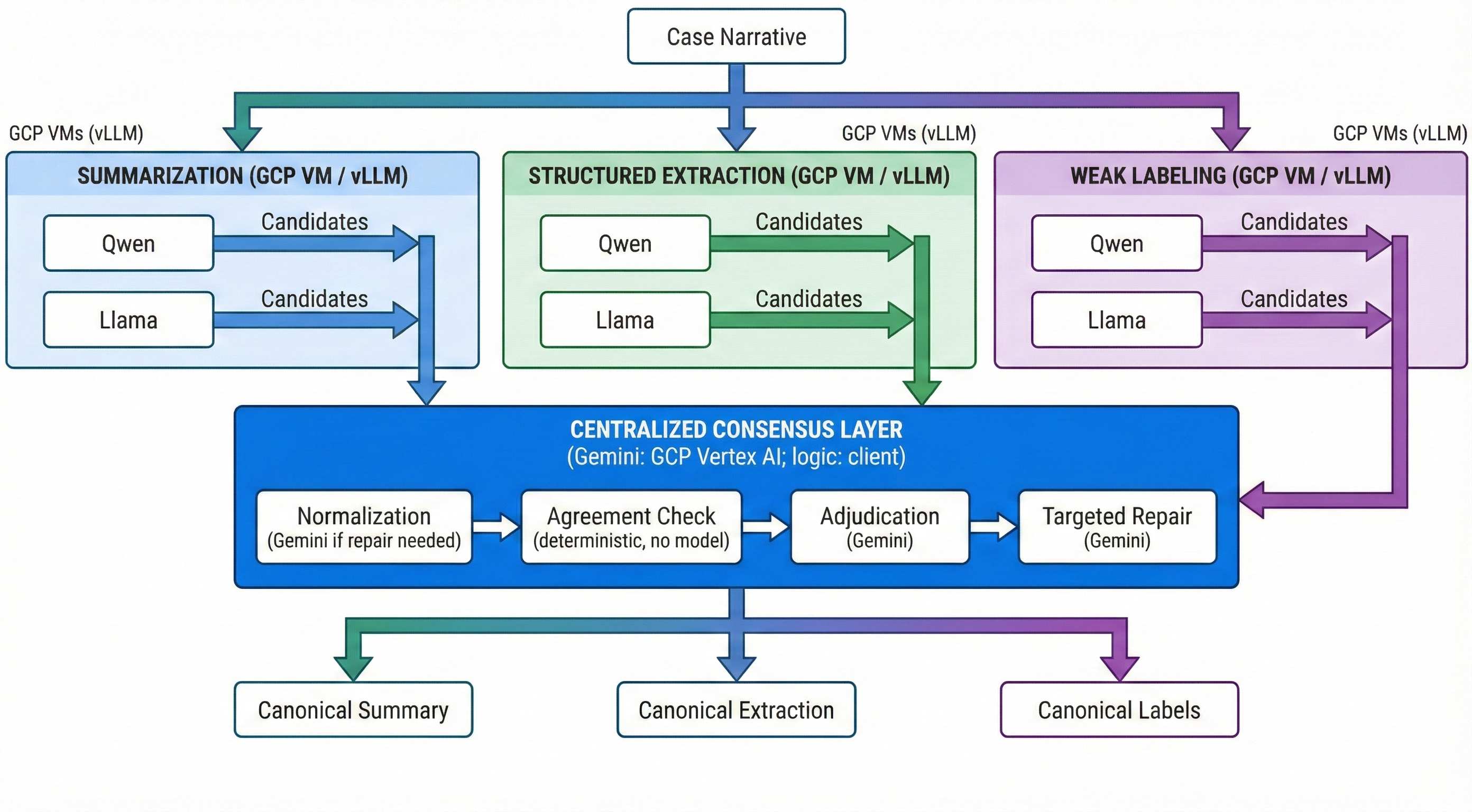}
\caption{Guardian LLM Pipeline Overview and Consensus Routing} \label{fig2}
\end{figure}
\subsection{Centralized Consensus Layer}
The consensus engine (Figure 3) is the primary reliability mechanism in Guardian Core. Its purpose is not merely to “choose a better answer,” but to enforce invariants: schema conformity, factual supportability relative to candidates, deterministic structure, and controlled behavior under disagreement. 
\begin{figure}
\includegraphics[width=\textwidth]{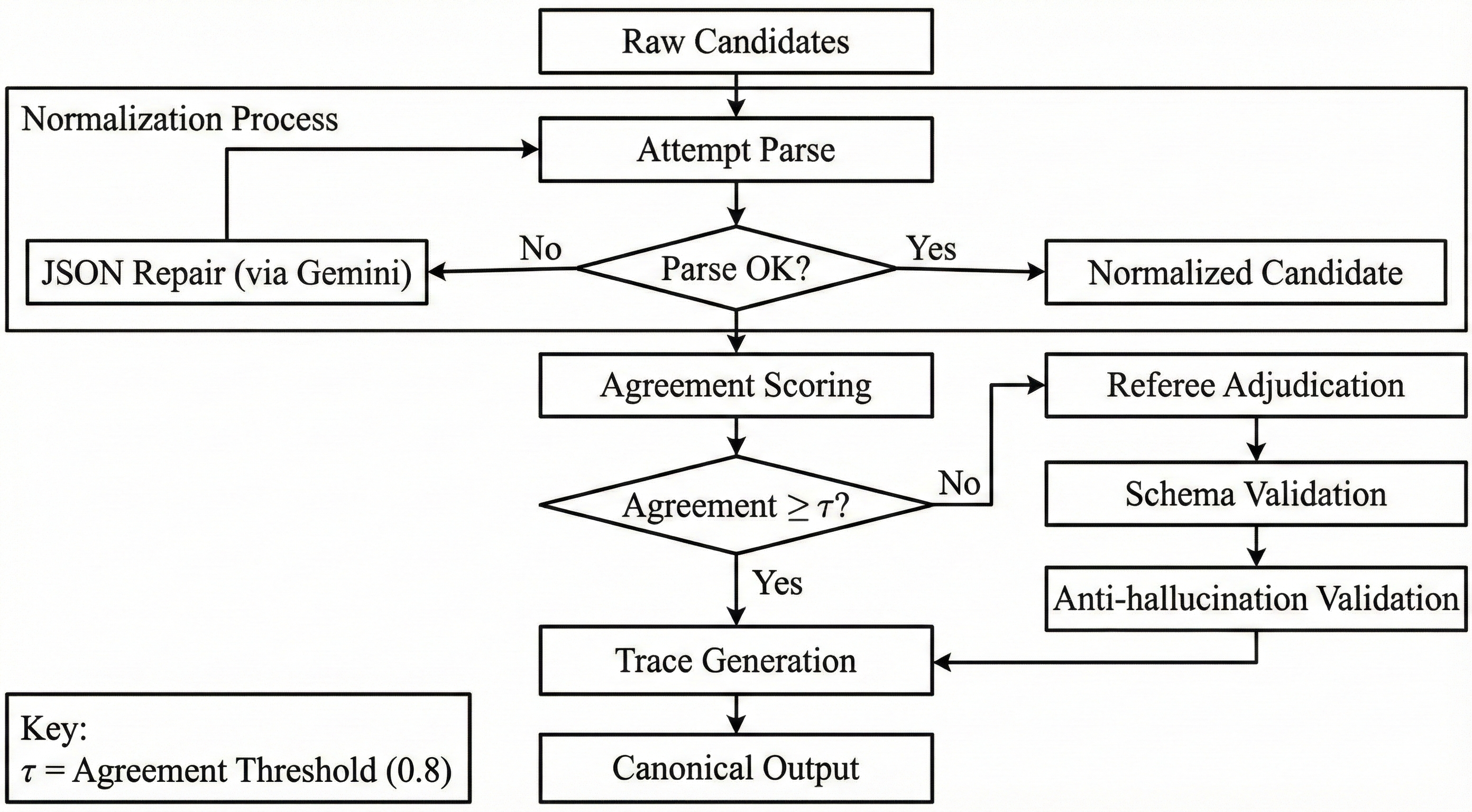}
\caption{Centralized Consensus Mechanism and Conflict Resolution Workflow} \label{fig3}
\end{figure}

The consensus process begins with normalization, which transforms each candidate into a comparable representation. For summaries, normalization enforces a fixed bullet structure and strips formatting idiosyncrasies so that agreement can be measured meaningfully. For extraction, normalization parses JSON using multiple strategies and coerces fields into schema-defined types, inserting empty defaults for missing required fields. For weak labels, normalization maps free-form terms to a closed label set, constraining movement and risk to valid categories and standardizing rationales.

Following normalization, the engine performs agreement scoring. Agreement is computed at the field level rather than solely at the surface-text level, because textual paraphrases can mask substantive discrepancies. For summaries, agreement is measured by comparing extracted informational slots such as subject identity, last-seen location and time, vehicle references, and movement cues, using token overlap and constrained similarity thresholds. For extraction, agreement compares key schema fields and treats lists as order-invariant sets where appropriate. This strategy is motivated by the observation that stable downstream analytics depend on stable structured fields, not on stylistic similarity.


The consensus layer also supports targeted repair when parsing fails. If extraction candidates contain malformed JSON, the system attempts recovery through deterministic extraction of JSON-like substrings and, only if necessary, invokes a repair prompt that instructs a model to return valid JSON matching the schema. Even in repair mode, the system validates repaired outputs and constrains them to the schema’s required fields, reflecting the broader practice in unstructured document pipelines of combining statistical methods with rule-based normalization and strict validation \cite{Mahadevkar2024}.

Finally, it emits trace artifacts that record whether the referee was called, which agreement thresholds were met, and which fields were repaired or reverted. These traces are crucial for auditability and for iterative improvement of prompts, validators, and downstream evaluation metrics, consistent with calls for transparent and responsible AI deployment \cite{Floridi2019,ICRC2025}.
\subsection{Backend Abstraction}
The backend module isolates the pipeline from the heterogeneity of model providers and deployment modes. 
The backend layer implements three main responsibilities: transport, resiliency, and policy. 

Transport standardizes how prompts are packaged and sent, including message formats for chat-style endpoints and parameter normalization. 
Resiliency encompasses retry policies with exponential backoff and jitter for transient failures, as well as strict timeouts to prevent indefinite blocking, a critical requirement in time-sensitive investigative workflows. Policy covers rate limiting and caching decisions that protect both cloud quotas and local GPU resources. 


\subsection{Orchestration and Concurrency}
The orchestrator module coordinates parallel execution across models and tasks, managing deadlines, concurrency limits, caching interactions, and the handoff into consensus. Orchestration is central in a consensus-first design because each case triggers multiple model calls per task; without careful scheduling, the system can overload compute resources, exceed rate limits, or create cascading delays that undermine operational utility. Guardian therefore treats orchestration as both a performance layer and a reliability layer.


In addition to concurrency, orchestration implements deadline management at the case and stage level. Each task receives a deadline budget, and the orchestrator continually checks remaining time before escalating to more expensive operations such as referee adjudication. If time is insufficient, the system returns the best validated candidate available rather than attempting a late referee call that could time out and yield no result. This time-aware behavior supports the investigative reality that timely, conservative outputs are often more valuable than delayed, potentially more refined outputs.

Orchestration also integrates caching to reduce redundant computation. When a case has been processed previously, the orchestrator can return cached results for a given stage, provided the caller has not requested a forced recomputation. 

Finally, the orchestrator defines the precise ordering by which candidate generation, normalization, consensus adjudication, validation, and persistence occur. This ordering matters because downstream analytics, i.e., clustering, mobility forecasting, and search-zone generation, depend on the stability and completeness of extracted fields \cite{Ester1996,Jiang2022}.

\subsection{Zone QA}
Zone QA (Quality Assurance) module extends the Guardian pipeline into the search-zone (a specific, defined, and delineated geographic area) domain by applying LLM-assisted plausibility scoring to candidate zones and reweighting zone priorities under an explicit reward configuration. 
Operationally, this module loads cases that include zone candidates and their associated metadata, optionally loads auxiliary zone scores (such as normalized RL scores), and processes cases in batches to compute plausibility assessments. 

The Zone QA output is not a free-form narrative judgment; it is a constrained score that is mapped from validated label outputs and then combined with existing zone signals using a transparent formula. The recomputation function explicitly weights original zone priority, LLM-derived plausibility, zone size penalties, and optional RL-derived scores, producing a bounded priority via a sigmoid transformation. 

This module is also designed to be safe under imperfect information. When the labeler fails or returns low-confidence results, Zone QA defaults plausibility to a neutral score rather than producing extreme penalties or boosts. This conservative behavior reflects ethical guidance for minimizing harm when automated systems operate on sensitive data, especially where false confidence could misdirect search resources \cite{Floridi2019,ICRC2025}. 

\section{LLM Prompting System and Template Governance}

\begin{figure}
\includegraphics[width=\textwidth]{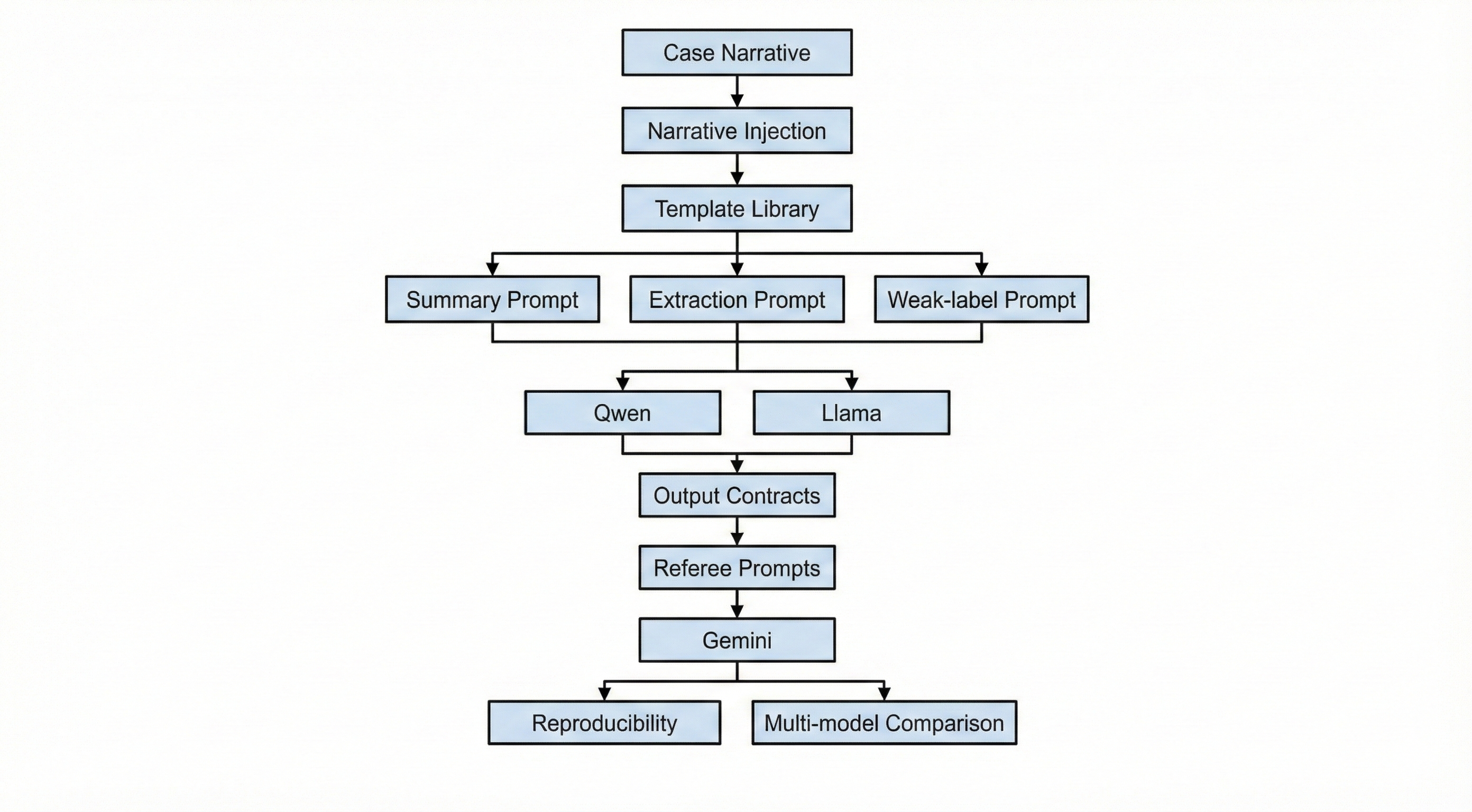}
\caption{Prompt Governance and Template-Based LLM Interaction} \label{fig3}
\end{figure}
The large language model (LLM) prompting architecture in Guardian is intentionally designed as a reliability-oriented mechanism rather than a mere usability or abstraction layer. This design choice aligns with the prevailing research consensus that language models exhibit superior performance when assigned narrowly scoped, well-specified roles governed by explicit output contracts and constraints \cite{Chau2002,Chen2024,Li2025}.

In Guardian, a prompt is treated as a first-class system artifact: it is the explicit, inspectable interface through which an investigative narrative is transformed into a bounded task request for an LLM backend. This framing is essential because missing-person investigations depend on converting noisy, incomplete narratives into structured, auditable intelligence under time pressure \cite{Solaiman2022,RuizReyes2025}. 

To achieve this, Guardian distinguishes three operational prompts: task prompts, consensus prompts, and format-guard prompts. 
Task prompts generate primary artifacts such as investigator summaries, schema-aligned entity extractions, and weak labels for movement and risk (Figure 4). Consensus prompts (or referee prompts) are invoked only when candidate outputs disagree or violate structural requirements, triggering adjudication, tie-breaking, or repair. Format-guard prompts are not a separate execution stage but a recurring design pattern in which prompts embed explicit contracts, such as “return JSON only,” fixed key sets, enumerated label spaces, and constrained bullet schemas. These ensure that outputs remain machine-actionable, comparable across models, and stable across repeated runs \cite{Mahadevkar2024}.

Beyond the orchestrator-driven pathway, Guardian includes role-specific local modules with prompts optimized for robustness in single-model contexts and for model-specific chat formats. The local summarizer prompt is intentionally minimal and imperative, requiring exactly five short bullets and forbidding commentary. This strict prompt contract, combined with deterministic post-processing, extracting bullet-like lines and enforcing a hard stop. This ensures stable summaries even when models emit extraneous text. The extractor module supports multiple prompt styles aligned with different reliability strategies.

Weak-labeling prompts adopt a similar dual strategy of explicit constraints and model-aligned formatting. The weak-labeler module includes a chat-formatted strict JSON prompt with explicit system and user role headers and an exact JSON example. 

The consensus layer introduces a distinct family of prompts whose purpose is not to generate primary artifacts from narratives but to reconcile candidate outputs produced by other models under strict constraints. These prompts fall into three principal categories: JSON repair prompts, referee adjudication prompts, and explicit tie-break prompts. Repair prompts are invoked only after deterministic parsing and recovery strategies fail.
Referee adjudication prompts are used when multiple candidates are structurally valid yet disagree on material fields. These prompts instruct the adjudicator to select between candidate values or merge compatible ones while explicitly prohibiting invention of facts not present in the candidates. 
Tie-break prompts further narrow the adjudicator’s scope by focusing on a single contested field and requiring JSON-only output with enumerated labels and a brief rationale. Summarization referee prompts adopt an additional efficiency strategy by using compact narrative prefixes and fixed five-line bullet schemas, reserving token budget for structured output while maintaining a stable contract for normalization.


Taken together, the prompting system in Guardian is best understood as an ecosystem of governed templates, controlled construction sites, and role-specific constraints that collectively enable deterministic validation and multi-model consensus. 

\section{QLoRA-Based Fine-Tuning Integration}
Guardian integrates QLoRA-based fine-tuning to improve role-specific performance while preserving scalability and multi-model flexibility. It enables role specialization by updating only a small set of low-rank adapter parameters on top of a quantized base model, preserving much of the base model’s general language competence while making training feasible on modest hardware \cite{Dettmers2023}. The fine-tuning workflow is designed around curated training inputs that reflect Guardian’s operational tasks. 

A key design decision is that fine-tuned models are integrated as interchangeable backends behind the same interface. This allows the pipeline to treat “fine-tuned Qwen extractor” and “fine-tuned Llama extractor” as peers generating candidates that are subsequently adjudicated by the consensus LLM. Such integration preserves the core principle that reliability is achieved through consensus, while fine-tuning improves candidate quality and reduces the burden on repair and re-ranking mechanisms. This is particularly consistent with cautionary findings that LLMs’ evaluation and judging capabilities can be limited. Accordingly, Guardian uses fine-tuned specialists to produce better candidates and uses structured consensus and validation to decide what is safe to accept \cite{Li2025}.

\section{Evaluation}
We have deployed Guardian as a distributed Google Cloud configuration consisting of three separate GPU virtual machines (VMs), each dedicated to a single task role: an extractor VM, a summarizer VM, and a weak-labeler VM, respectively. On each VM, two models run concurrently, a Qwen2.5-3B-Instruct model and a Llama-3.2-3B-Instruct model. Each model is served via a Dockerized vLLM server exposing an 
OpenAI-compatible API. Operationally, the Qwen server binds to port 8001 on each VM and the Llama server binds to port 8002 on each VM, producing six total inference servers running in parallel across the three machines. This separation of concerns allows each VM to be tuned for its workload while supporting consistent invocation semantics across roles.
The outputs of the individual models are integrated within a consensus layer implemented using Gemini 2.5 Flash/Pro. This layer is employed for consensus formation, adjudication among model outputs, and automated JSON structure repair (via Vertex AI). It resolves conflicts between predictions generated by Qwen and Llama, corrects malformed or nonconformant JSON responses, and can optionally assist in the normalization phase. It is not involved in initial summarization, information extraction, or weak-label generation. Figures 5 and 6 present representative JSON output excerpts produced by the Llama LLM and by the Gemini-based consensus LLM, respectively.

\begin{figure}
\includegraphics[scale=0.50]{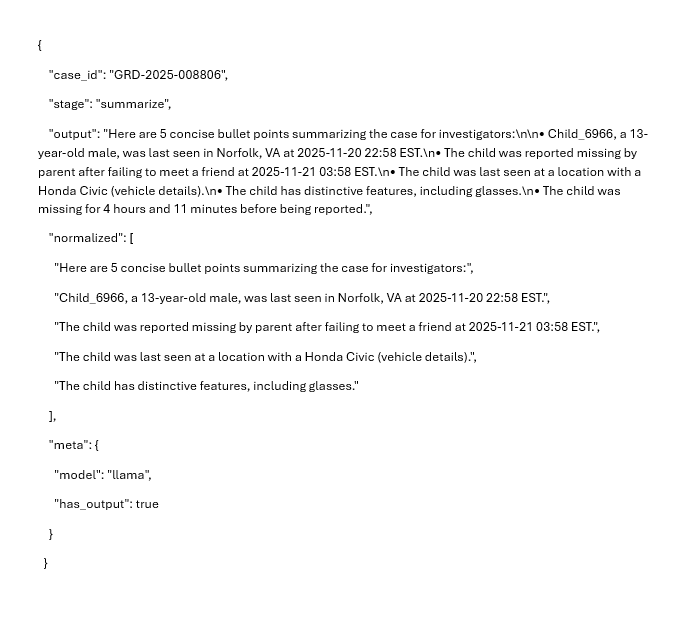}
\caption{Excerpt from the Llama LLM JSON Output} \label{fig5}
\end{figure}
\begin{figure}
\includegraphics[scale=0.5]{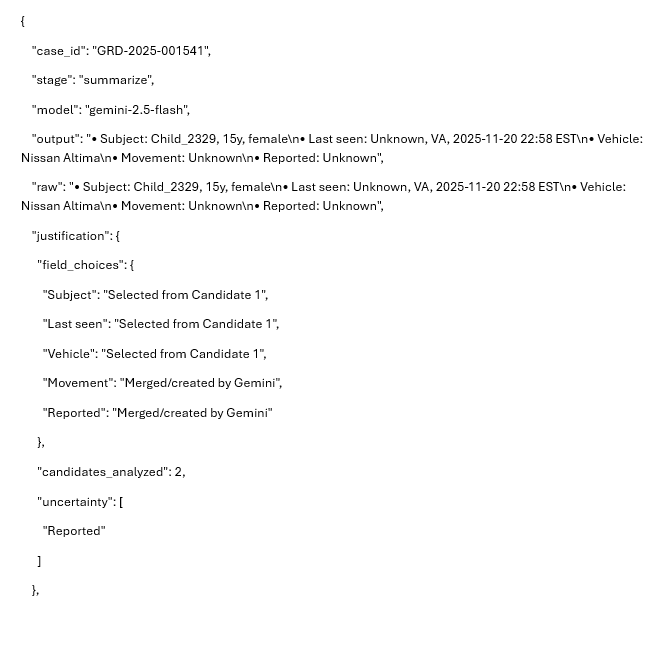}
\caption{Excerpt from the Gemini Consensus LLM JSON Output} \label{fig5}
\end{figure}
In the present study, reliability is assessed operationally rather than through a single aggregate metric. At the structural level, we examine whether outputs are parseable, schema-aligned, and repairable when malformed. At the factual level, we compare extraction and weak-label outputs against structured synthetic case ground truth where available. At the system level, we consider agreement across candidate models, the need for referee intervention, fallback behavior, and the extent to which the pipeline continues to produce stable, auditable outputs under disagreement and failure.

Guardian’s evaluation emphasizes reliability, structural validity, and failure-mode behavior, rather than benchmark-style predictive accuracy. This focus reflects both the safety-critical nature of missing-child investigations and the well-documented lack of complete ground truth in real-world disappearance cases, particularly during early-stage response \cite{FBI2014, RuizReyes2025,Solaiman2022}. In such contexts, prior work argues that the primary risk is not marginal predictive error but the propagation of unstable or unsupported inferences into operational decision-making \cite{Allen1998,Budowle2024PrivacyFGG}. Accordingly, the evaluation is framed as a qualitative, diagnostic analysis of pipeline behavior under realistic operating conditions, grounded in direct inspection of live model outputs produced during system execution.

The evaluation corpus consists of synthetic and semi-structured missing-child case narratives processed end-to-end through the Guardian LLM Pipeline. Synthetic data is used to introduce controlled variation while avoiding exposure of sensitive real cases. This is consistent with established practices for developing and stress-testing AI systems in domains with privacy and availability constraints \cite{Nikolenko2019,Sun2023SyntheticHealthData}. For each case, the pipeline produces three primary artifacts: a concise investigator-facing summary, a schema-aligned structured extraction, and a weak-label output capturing movement and risk signals. Outputs are examined at both the candidate level (individual model responses) and the consensus level (final canonical outputs after normalization, validation, and adjudication). This two-level inspection aligns with prior research showing that understanding how errors arise in intermediate representations is essential for reliable downstream analytics in unstructured document pipelines \cite{Chau2002,Mahadevkar2024}.

Inspection of raw single-model outputs reveals several recurring failure modes that directly motivate Guardian’s architecture. In weak labeling, individual models frequently exhibit overconfident classification, assigning high-risk or long-range movement labels even when narrative evidence is ambiguous or incomplete. This tendency is consistent with recent findings that LLMs often produce confident categorical outputs despite uncertainty, particularly when asked to act as implicit judges without explicit constraints \cite{Chen2024,Li2025}. 

In summarization, models often generate readable but speculative statements, such as inferred offender intent or implied vehicle usage not explicitly supported by the narrative, a pattern that echoes longstanding concerns about narrative interpretation in child-abduction contexts where unsupported assumptions can distort risk perception \cite{Allen1998,Miller2008}. 
In extraction tasks, malformed or partially invalid JSON outputs appear regularly, especially when narratives are long or noisy, reflecting broader challenges in unstructured document analysis where schema compliance cannot be assumed without explicit validation \cite{bird2009nltk,Mahadevkar2024}.

When Guardian’s normalization and consensus mechanisms are applied, outputs become systematically more conservative and structurally consistent. Normalization enforces closed vocabularies, fixed schemas, and deterministic parsing rules, collapsing free-form language into representations suitable for downstream processing and analysis \cite{Aggarwal2015}. The centralized consensus layer further reduces volatility by comparing candidate outputs across models, suppressing unsupported fields, and reconciling disagreements only within the bounds of observed evidence. This approach operationalizes insights from weak supervision and multi-source labeling research, which show that aggregating multiple noisy signals under explicit constraints yields more reliable supervision than trusting any single source \cite{Ratner2017,Zeng2022}. In cases where models disagree, consensus outputs consistently favor interpretations that are explicitly grounded in the narrative, rather than those implied by model confidence alone, aligning with recent critiques of unconstrained LLM self-evaluation and judging \cite{Li2025}.

Although this evaluation does not claim quantitative improvements in investigative outcomes or recovery rates, it demonstrates that Guardian reliably converts unstructured, narrative-driven inputs into auditable, schema-aligned artifacts that remain stable across repeated runs. These properties are essential for downstream geospatial modeling, clustering, and search-zone generation, which depend on consistent structured inputs rather than stylistically variable text 
\cite{Ester1996,Jiang2022,Lyu2025}. In this sense, the evaluation supports the central thesis of this work: in missing-child and other safety-adjacent domains, reliability must be enforced at the system level through validation, normalization, and consensus, rather than assumed from individual model outputs or generative confidence \cite{Floridi2019,ICRC2025}.

\section{Discussion}
Guardian’s architecture treats reliability as a systems property. The primary contribution is not any single model’s accuracy, but the end-to-end design in which multi-model candidate generation is forced through validation and consensus before outputs are accepted. This approach is particularly appropriate for missing-person and child-safety contexts, where narratives are incomplete and high-stakes decisions must remain auditable. The consensus layer operationalizes a pragmatic view of LLMs: they are powerful pattern extractors and summarizers, but they should be constrained, cross-checked, and integrated as components within a larger verification pipeline rather than treated as authoritative. This is consistent with research arguing that LLMs are often most effective as structured annotators that generate useful supervision and intermediate artifacts, especially when human oversight and validation mechanisms are present \cite{Chen2024}.

The QLoRA integration demonstrates how parameter-efficient fine-tuning strengthens candidate quality without undermining scalability.By adapting fewer than 1\% of parameters, Guardian can train role-specific specialist models while preserving the feasibility of inference across multiple models and roles. The observed training dynamics—specifically, monotonic reductions in loss, stabilizing gradient norms, and well-behaved learning-rate schedules—indicate that the models are internalizing the constrained behaviors required for operational deployment. Within a consensus-first framework, these improvements exhibit multiplicative downstream effects: higher-quality candidate outputs reduce the need for post hoc repair, diminish ambiguity during adjudication, and increase the proportion of instances in which models concur, thereby enhancing system stability and lowering overall computational expenditure.

The reproducibility and deployment design highlights a practical lesson: multi-model consensus requires not only algorithmic strategies but also robust infrastructure. By distributing roles across dedicated GPU VMs, serving models through standardized APIs, and tunneling endpoints securely to the coordinator, Guardian makes multi-model inference both scalable and operationally manageable. This infrastructure approach supports privacy and safety expectations by minimizing exposure and enabling controlled access, consistent with broader concerns about responsible AI use in sensitive investigative domains \cite{Floridi2019, ICRC2025}.

\subsection{Limitations}
We should also note some of the primary limitations of our system. First, Guardian’s consensus strategy increases computational cost relative to single-model pipelines because each task is replicated across multiple models by design. While orchestration and parallel execution mitigate latency, resource consumption remains a tradeoff. This design is chosen intentionally because operational stability is prioritized.

A second limitation concerns ground-truth scarcity in missing-person contexts. Weak labels and extraction outputs can be validated structurally, but semantic correctness can be difficult to confirm without authoritative datasets. Guardian therefore emphasizes auditable traces and conservative acceptance policies, but empirical evaluation will remain constrained by data availability and the sensitivity of investigative records \cite{Solaiman2022, RuizReyes2025}. 

A third limitation is that LLMs, even when fine-tuned, may be unreliable at complex spatial reasoning tasks without explicit geospatial algorithms. Guardian addresses this by restricting LLM roles to extraction, summarization, and weak labeling, and by validating geospatial outputs through Zone QA rather than trusting generative reasoning \cite{Xu2025,Truong2025}. 

\section{Conclusion}
This paper described the Guardian LLM Pipeline, a consensus-routed, multi-model architecture designed to produce reliable, schema-aligned outputs for missing-person intelligence workflows. The system’s primary entry point module coordinates end-to-end execution, while orchestrator module enables scalable multi-model parallelism and the backend module standardizes heterogeneous providers behind consistent interfaces. Reliability is centered in the consensus module, which compares multiple model outputs, resolves disagreements through structured conflict-resolution strategies, and applies repair and verification to produce stable canonical results. Robustness is strengthened by the Zone QA  module, which detects and corrects zone-level and structural issues before and after consensus evaluation. QLoRA-based fine-tuning improves candidate quality using curated datasets composed of synthetic cases, real cases, and research-derived corpora, while updating less than 1\% of model parameters to preserve scalability. 
Taken together, these design choices support a pragmatic view of reliability for safety-adjacent LLM systems: dependable behavior arises from well-bounded roles, explicit checking, and centralized agreement, rather than from trusting any single model on its own.

\bibliographystyle{splncs04}
\bibliography{references}

@book{Aggarwal2015,
  author    = {Aggarwal, Charu C.},
  title     = {Data Mining: The Textbook},
  publisher = {Springer},
  year      = {2015}
}

@article{Allen1998,
  author  = {Allen, Edward E.},
  title   = {Keeping Children Safe: Rhetoric and Reality},
  journal = {Juvenile Justice Journal},
  volume  = {5},
  number  = {1},
  year    = {1998},
  month   = may,
  note    = {U.S. Department of Justice, Office of Juvenile Justice and Delinquency Prevention}
}

@article{Chau2002,
  author  = {Chau, Michael and Xu, Jennifer J. and Chen, Hsinchun},
  title   = {Extracting Meaningful Entities from Police Narrative Reports},
  journal = {Journal of the American Society for Information Science and Technology},
  volume  = {53},
  number  = {11},
  pages   = {984--995},
  year    = {2002}
}

@inproceedings{Chen2024,
  author    = {Chen, Rui and Qin, Chao and Jiang, Wei and Choi, Dongwon},
  title     = {Is a Large Language Model a Good Annotator for Event Extraction?},
  booktitle = {Proceedings of the AAAI Conference on Artificial Intelligence (AAAI-24)},
  pages     = {17772--17780},
  year      = {2024}
}

@misc{Dettmers2023,
  author = {Dettmers, Tim and Pagnoni, Artidoro and Holtzman, Ari and Zettlemoyer, Luke},
  title  = {QLoRA: Efficient Finetuning of Quantized Large Language Models},
  note   = {arXiv preprint arXiv:2305.14314},
  year   = {2023}
}

@misc{Ewers2024,
  author = {Ewers, Richard and Anderson, James and Thomson, David},
  title  = {Agent-Based Predictive Probability Density Mapping for Search and Rescue},
  year   = {2024},
  note   = {Manuscript}
}

@book{FBI2014,
  author    = {{Federal Bureau of Investigation}},
  title     = {Child Abduction Response Plan: An Investigative Guide},
  edition   = {3},
  publisher = {U.S. Department of Justice},
  year      = {2014}
}

@article{Floridi2019,
  author  = {Floridi, Luciano and Cowls, Josh},
  title   = {A Unified Framework of Five Principles for AI in Society},
  journal = {Harvard Data Science Review},
  volume  = {1},
  number  = {1},
  year    = {2019}
}

@article{Hashimoto2022,
  author  = {Hashimoto, Aiko and Heintzman, Logan and Koester, Robert and Abaid, Nicole},
  title   = {An Agent-Based Model Reveals Lost Person Behavior Based on Data from Wilderness Search and Rescue},
  journal = {Scientific Reports},
  volume  = {12},
  pages   = {5873},
  year    = {2022}
}

@techreport{ICRC2025,
  author      = {{International Committee of the Red Cross}},
  title       = {Balancing Risks and Opportunities: New Technologies and the Search for Missing People},
  institution = {ICRC},
  year        = {2025}
}

@article{Jiang2022,
  author  = {Jiang, Wei and Luo, Jun},
  title   = {Graph Neural Network for Traffic Forecasting: A Survey},
  journal = {Expert Systems with Applications},
  volume  = {207},
  pages   = {117921},
  year    = {2022}
}

@inproceedings{Ester1996,
  author    = {Ester, Martin and Kriegel, Hans{-}Peter and Sander, J{\"o}rg and Xu, Xiaowei},
  title     = {A Density-Based Algorithm for Discovering Clusters in Large Spatial Databases with Noise},
  booktitle = {Proceedings of the Second International Conference on Knowledge Discovery and Data Mining (KDD-96)},
  pages     = {226--231},
  publisher = {AAAI Press},
  year      = {1996}
}

@book{Cormen2022,
  author    = {Cormen, Thomas H. and Leiserson, Charles E. and Rivest, Ronald L. and Stein, Clifford},
  title     = {Introduction to Algorithms},
  edition   = {4},
  publisher = {The MIT Press},
  year      = {2022}
}

@book{bird2009nltk,
  author    = {Bird, Steven and Klein, Ewan and Loper, Edward},
  title     = {Natural Language Processing with Python},
  publisher = {O'Reilly Media},
  year      = {2009}
}

@misc{Li2025,
  author = {Li, Tingting and Qin, Yifan and Sheng, Olivia R. L.},
  title  = {A Multi-Task Evaluation of LLMs' Processing of Academic Text Input},
  note   = {arXiv preprint arXiv:2508.11779},
  year   = {2025}
}

@inproceedings{Lyu2025,
  author    = {Lyu, Feng},
  title     = {Evaluating the Evaluation Matrices: Integrating Spatial Assessment in Geospatial AI Model Training and Evaluation},
  booktitle = {I-GUIDE Forum 2025},
  year      = {2025}
}

@article{Mahadevkar2024,
  author  = {Mahadevkar, S. V. and Patil, S. and Kotecha, K. and Soong, L. W. and Choudhury, T.},
  title   = {Exploring AI-Driven Approaches for Unstructured Document Analysis and Future Horizons},
  journal = {Journal of Big Data},
  volume  = {11},
  pages   = {92},
  year    = {2024}
}

@article{Miller2008,
  author  = {Miller, Jody M. and Kurlychek, Megan and Hansen, John A. and Wilson, Kyle},
  title   = {Examining Child Abduction by Offender Type Patterns},
  journal = {Justice Quarterly},
  volume  = {25},
  number  = {3},
  pages   = {523--543},
  year    = {2008}
}

@misc{Nikolenko2019,
  author = {Nikolenko, Sergey I.},
  title  = {Synthetic Data for Deep Learning},
  note   = {arXiv preprint arXiv:1909.11512},
  year   = {2019}
}

@article{Papic2024,
  author  = {Papi{\'c}, Vedran and {\v S}ari{\'c} Gudelj, Ana and Milan, Ante and Mili{\v c}evi{\'c}, Marko},
  title   = {Person Mobility Algorithm and Geographic Information System for Search and Rescue Missions Planning},
  journal = {Remote Sensing},
  volume  = {16},
  number  = {4},
  pages   = {670},
  year    = {2024}
}

@article{Ratner2017,
  author  = {Ratner, Alexander and Bach, Stephen H. and Ehrenberg, Henry and Fries, Jason and Wu, Sen and R{\'e}, Christopher},
  title   = {Snorkel: Rapid Training Data Creation with Weak Supervision},
  journal = {Proceedings of the VLDB Endowment},
  volume  = {11},
  number  = {3},
  pages   = {269--282},
  year    = {2017}
}

@article{RuizReyes2025,
  author  = {Ruiz Reyes, Jorge and Congram, Derek and Sirbu, Radu A. and Floridi, Luciano},
  title   = {Where Are They? A Review of Statistical Techniques and Data Analysis to Support the Search for Missing Persons and the New Field of Data-Based Disappearance Analysis},
  journal = {Forensic Science International},
  volume  = {376},
  pages   = {112582},
  year    = {2025}
}

@article{Solaiman2022,
  author  = {Solaiman, K. M. A. and Sun, Tianyi and Nesen, Andrey and Bhargava, Bharat and Stonebraker, Michael},
  title   = {Applying Machine Learning and Data Fusion to the ``Missing Person'' Problem},
  journal = {IEEE Computer},
  volume  = {55},
  number  = {6},
  pages   = {40--55},
  year    = {2022}
}

@misc{Truong2025,
  author = {Truong, Tuan H. and Lau, Jey Han and Qi, Jianzhong},
  title  = {Understanding the Geospatial Reasoning Capabilities of LLMs: A Trajectory Recovery Perspective},
  note   = {arXiv preprint arXiv:2510.01639},
  year   = {2025}
}

@misc{Xu2025,
  author = {Xu, Lei and Zhao, Shiyu and Lin, Qiang and Chen, Liang and Luo, Qiang and Wu, Shuai and Ye, Xiang and Feng, Hao and Du, Zongmin},
  title  = {Evaluating Large Language Models on Spatial Tasks: A Multi-Task Benchmarking Study},
  note   = {arXiv preprint arXiv:2408.14438},
  year   = {2025}
}

@inproceedings{Zeng2022,
  author    = {Zeng, Zihan and Ni, Wei and Fang, Tianyi and Li, Xiaonan and Zhao, Xin and Song, Yangqiu},
  title     = {Weakly Supervised Text Classification Using Supervision Signals from a Language Model},
  booktitle = {Findings of NAACL 2022},
  pages     = {2295--2305},
  year      = {2022}
}

@article{Budowle2024PrivacyFGG,
  author    = {Budowle, Bruce and Baker, L. and Sajantila, Antti and Mittelman, K. and Mittelman, David},
  title     = {Prioritizing privacy and presentation of supportable hypothesis testing in forensic genetic genealogy investigations},
  journal   = {BioTechniques},
  volume    = {76},
  number    = {9},
  pages     = {425--431},
  year      = {2024},
}

@article{Sun2023SyntheticHealthData,
  author    = {Sun, C. and van Soest, J. and Dumontier, M.},
  title     = {Generating synthetic personal health data using conditional generative adversarial networks combining with differential privacy},
  journal   = {Journal of Biomedical Informatics},
  volume    = {143},
  pages     = {104404},
  year      = {2023},
  doi       = {10.1016/j.jbi.2023.104404}
}
%




\end{document}